\pdfoutput=1

\documentclass[11pt]{article}

\usepackage{ACL2023}

\usepackage{times}
\usepackage{latexsym}
\usepackage{booktabs}
\usepackage{rotating}
\usepackage{multirow}

\usepackage{CJKutf8}

\usepackage[T1]{fontenc}

\usepackage[utf8]{inputenc}

\usepackage{microtype}

\usepackage{inconsolata}

\title{tmn at SemEval-2023 Task 9: Multilingual Tweet Intimacy Detection using XLM-T, Google Translate, and Ensemble Learning}

\author{Anna Glazkova \\
  University of Tyumen \\
  \texttt{a.v.glazkova@utmn.ru} }

\begin{document}
\maketitle
\begin{abstract}
The paper describes a transformer-based system designed for SemEval-2023 Task 9: Multilingual Tweet Intimacy Analysis. The purpose of the task was to predict the intimacy of tweets in a range from 1 (not intimate at all) to 5 (very intimate). The official training set for the competition consisted of tweets in six languages (English, Spanish, Italian, Portuguese, French, and Chinese). The test set included the given six languages as well as external data with four languages not presented in the training set (Hindi, Arabic, Dutch, and Korean). We presented a solution based on an ensemble of XLM-T, a multilingual RoBERTa model adapted to the Twitter domain. To improve the performance of unseen languages, each tweet was supplemented by its English translation. We explored the effectiveness of translated data for the languages seen in fine-tuning compared to unseen languages and estimated strategies for using translated data in transformer-based models. Our solution ranked 4th on the leaderboard while achieving an overall Pearson's r of 0.599 over the test set. The proposed system improves up to 0.088 Pearson's r over a score averaged across all 45 submissions.


\end{abstract}

\section{Introduction}

Intimacy is a significant social aspect of language, which helps to explore existing social norms in various contexts \cite{pei2020quantifying}. The concept of intimacy describes how an individual relates to his addressee in their perceived interdependence, warmth, and willingness to personally share \cite{perlman1987development}. An automatic evaluation of intimacy in language provides us with a clearer picture of social interactions and different linguistic strategies.

This paper describes a system developed for the SemEval-2023 Task 9:
Multilingual Tweet Intimacy Analysis \cite{pei2022semeval}. The purpose of the task was to predict the intimacy of tweets. The test set contained tweets in the languages presented (seen) and not presented (unseen) in the training set. The main challenge of the task was the zero-shot intimacy prediction performance for unseen languages.

\begin{figure}
\centerline{\includegraphics[scale=0.35]{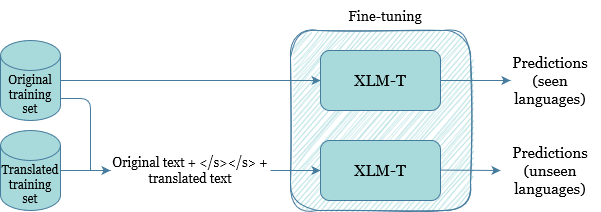}}
\caption{High-level overview of our approach.}
\label{fig}
\end{figure}

Inspired by the recent success of transformer-based models in multilingual tasks \cite{malmasi-etal-2022-semeval,tayyar-madabushi-etal-2022-semeval}, we evaluate the performance of BERT (Bidirectional Encoder Representations from Transformers) \cite{devlin-etal-2019-bert} and its modifications. In our experiments, we investigated the impact of translation to improve the zero-shot performance. We have found that the joint use of original and translated texts increases the model performance for unseen languages in comparison with the separate use of original or translated data.

The paper is organized as follows. Section \ref{sec2} presents the task and the dataset provided by the task organizers. Our methods are presented in Section \ref{sec3}. In Section \ref{sec4}, we describe the experimental setups that we used during the development phase of the competition. We discuss our results in Section \ref{sec5}. Finally, Section \ref{sec6} concludes this paper.



\section{Task Description}\label{sec2}

The task aims to predict the intimacy of tweets in ten languages. The organizers presented a training set containing tweets in six languages (English, Spanish, Italian, Portuguese, French, and Chinese) annotated with intimacy scores ranging between 1 and 5 where 1 indicates “not intimate at all” and 5 indicates “very intimate”. The model performance was evaluated on the test set consisting of tweets in the given six languages and the four unseen languages that were not included in the training data (Hindi, Arabic, Dutch, and Korean). The total amount of entries in the training set is 9,491. The size of the test set is 3,881. Pearson's r was utilized as an evaluation metric. The dataset statistics are presented in Table \ref{table:stat}. A sample of annotated tweets for English is shown in Appendix \ref{appendix}. Figure \ref{ris:image} demonstrates the intimacy distribution for the training set. 

\begin{table}[]
\centering
\small
\begin{tabular}{|lll|}
\hline
\multicolumn{1}{|l|}{Language} & \multicolumn{1}{l|}{Train} & \multicolumn{1}{l|}{Test} \\ \hline
\multicolumn{3}{|c|}{\textit{Amount (avg number of tokens)}} \\ \hline
\multicolumn{1}{|l|}{English} & \multicolumn{1}{c|}{1,587 (23.99)} & 396 (22.79) \\ \hline
\multicolumn{1}{|l|}{Spanish} & \multicolumn{1}{c|}{1,592 (20.68)} & 399 (20.05)\\ \hline
\multicolumn{1}{|l|}{Italian} & \multicolumn{1}{c|}{1,532 (21.48)} & 384 (20.71)\\ \hline
\multicolumn{1}{|l|}{Portuguese} & \multicolumn{1}{c|}{1,596 (19.66)} & 398 (20.69)\\ \hline
\multicolumn{1}{|l|}{French} & \multicolumn{1}{c|}{1,588 (19.36)} & 393 (20.43)\\ \hline
\multicolumn{1}{|l|}{Chinese} & \multicolumn{1}{c|}{1,596 (29.97)} & 400 (28.19)\\ \hline
\multicolumn{1}{|l|}{Hindi} & \multicolumn{1}{c|}{-} & 280 (21.22)\\ \hline
\multicolumn{1}{|l|}{Arabic} & \multicolumn{1}{c|}{-} & 407 (21.34)\\ \hline
\multicolumn{1}{|l|}{Dutch} & \multicolumn{1}{c|}{-} & 413 (19.6)\\ \hline
\multicolumn{1}{|l|}{Korean} & \multicolumn{1}{c|}{-} & 411 (19.37) \\ \hline
\end{tabular}
\caption{\label{font-table} Data statistics. The number of tokens is obtained using the tokenizer of XLM-T \cite{barbieri-etal-2022-xlm}.}
\label{table:stat}
\end{table}

\begin{figure}[]
\center{\includegraphics[width=0.8\linewidth]{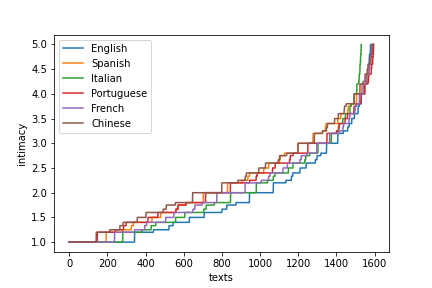}}
\caption{The distribution of intimacy scores for each language (training set).}
\label{ris:image}
\end{figure}

\section{System Overview}\label{sec3}

\subsection{Translated Tweets}

Previous studies have shown the effectiveness of the use of translated datasets for zero-shot learning \cite{schwenk-douze-2017-learning,eriguchi2018zero,ebrahimi-etal-2022-americasnli}. We used the public translation engine Google Translate\footnote{\url{https://translate.google.com/}} and the deep-translator Python tool\footnote{\url{https://github.com/nidhaloff/deep-translator}} to translate the training and test sets to English. We compared three strategies for representing input data: a) original dataset, i.e. texts are fed into the model in the same form as they are presented in the original dataset; b) translated dataset, i.e. all texts in all languages are replaced by their translations into English; c) joint dataset, each entry of which contains an original and translated text connected by a sequence of characters \textit{</s></s>}. For English tweets, the original and translated tweets match. The examples of input texts are given in Table \ref{ris:examples}.

\begin{table*}[]
\centering
\small
\begin{tabular}{|p{4cm}|p{4cm}|p{6cm}|}
\hline
Original & Translated & Original + translated \\ \hline
j’ai plus aucune force & I have no more strength & j’ai plus aucune force </s></s> I have no more strength \\ \hline
La mia prima stagione e, forse per questo, la mia favorita buon compleanno \#Reggina107 http & My first season and, perhaps for this reason, my favorite happy birthday \#Reggina107 http &  La mia prima stagione e, forse per questo, la mia favorita buon compleanno \#Reggina107 http </s></s> My first season and, perhaps for this reason, my favorite happy birthday \#Reggina107 http\\ \hline
@user Es normal cuando no se sale bien & @user It's normal when it doesn't work out &  @user Es normal cuando no se sale bien </s></s> @user It's normal when it doesn't work out\\ \hline
\begin{CJK*}{UTF8}{gbsn}
开学了 会更新的慢一点了 快万粉了 你们有什么想看的吗
\end{CJK*} http& The school has started, the update will be a bit slower, I am almost 10,000 fans, do you have anything you want to see http & \begin{CJK*}{UTF8}{gbsn}
开学了 会更新的慢一点了 快万粉了 你们有什么想看的吗
\end{CJK*} http </s></s> The school has started, the update will be a bit slower, I am almost 10,000 fans, do you have anything you want to see http\\ \hline
\end{tabular}
\caption{Strategies for input representation.}
\label{ris:examples}
\end{table*}

\subsection{Models}

We compared several pre-trained language models on the tweet intimacy prediction task:

\begin{itemize}
    \item BERT\footnote{\url{https://huggingface.co/bert-base-multilingual-cased}} \cite{devlin-etal-2019-bert}: BERT multilingual base model (cased), a multilingual version of BERT pre-trained on 104 languages.
    \item XLM-R\footnote{\url{https://huggingface.co/xlm-roberta-base}} \cite{DBLP:journals/corr/abs-1911-02116}: XLM-RoBERTa base, a model pre-trained on 2.5TB of filtered CommonCrawl data containing 100 languages. 
    \item XLM-T\footnote{\url{https://huggingface.co/cardiffnlp/twitter-xlm-roberta-base}} \cite{barbieri-etal-2022-xlm}: Twitter-XLM-Roberta base, a multilingual RoBERTa model trained over 200M tweets.
    \item RoBERTa\footnote{\url{https://huggingface.co/roberta-base}} \cite{liu2019roberta}: RoBERTa base model, a modification of BERT that is pre-trained using dynamic masking. Since RoBERTa is a monolingual model, we utilized it only using data translated into English.
\end{itemize}

All models were implemented using the Simple Transformers\footnote{\url{https://github.com/ThilinaRajapakse/simpletransformers}} library based on Transformers\footnote{\url{https://github.com/huggingface/transformers}} \cite{wolf-etal-2020-transformers} by HuggingFace. We fine-tuned each model for 3 epochs using a batch size of 8, a maximum sequence length of 128, and a learning rate of 4e-5.

The code we used to fine-tune models as well as the translations we made are presented on Github\footnote{\url{https://github.com/oldaandozerskaya/intimacy_tmn_semeval23_task9}}. 

\section{Experimental Setup}\label{sec4}

In this section, we describe the experimental setup utilized during the development phase of the competition. In this phase, the organizers provided participants with a training set containing 9,491 entries. We shuffled the provided data with a fixed random seed and divided it into training and validation subsets in a ratio of 70:30. Therefore, the training and validation subsets consisted of 6,643 and 2,848 entries respectively. To estimate the performance for unseen languages, we consistently excluded one language from the training set. Therefore, the model was fine-tuned on the five remaining languages and evaluated on the unseen language.

To evaluate the results during the development phase, we utilized two metrics. The first one was Pearson's r which was an official evaluation metric. The second metric we used was the mean squared error (MSE) which estimates the mean squared distances between actual and predicted values. 

\section{Results}\label{sec5}

\subsection{Development Phase}

We compared the performance of several pre-trained transformer-based models fine-tuned on the training subset formed during the development phase. The models were evaluated on the validation subset that did not participate in fine-tuning. All languages presented in the validation subset were also presented in the training subset. Therefore, the validation subset did not contain unseen languages.

The results of the comparison are presented in Table \ref{table:dev_all}. The best results in terms of MSE and Pearson's r were achieved by the XLM-T fine-tuned on the original dataset. Similar to the scores obtained by \citet{pei2022semeval}, our results showed the superiority of the XLM-T over the XLM-R and BERT for the task of multilingual tweet intimacy prediction. Fine-tuning on the translated dataset improves the results of BERT; however, the scores for XLM-R and XLM-T were worse in comparison with the results achieved while fine-tuning on the original dataset. The joint use of original and translated data leads to the highest Pearson's r score among the BERT models. In general, joint fine-tuning outperformed fine-tuning on translated data only but did not surpass the performance of the models fine-tuned on the original texts. The results in terms of MSE scores do not always correspond to the values of Pearson's r. In particular, the XLM-R demonstrated its best MSE and worst Pearson's r using translated dataset.

\begin{table}[]
\centering
\small
\begin{tabular}{|lll|}
\hline
\multicolumn{1}{|l|}{Model} & \multicolumn{1}{l|}{\textit{MSE}} & \textit{r} \\ \hline
\multicolumn{3}{|c|}{\textit{Original data}} \\ \hline
\multicolumn{1}{|l|}{BERT} & \multicolumn{1}{l|}{0.562} & 0.18 \\ \hline
\multicolumn{1}{|l|}{XLM-R} & \multicolumn{1}{l|}{0.499} & 0.517 \\ \hline
\multicolumn{1}{|l|}{XLM-T} & \multicolumn{1}{l|}{\textbf{0.437}} & \textbf{0.572} \\ \hline
\multicolumn{3}{|c|}{\textit{Translated data}} \\ \hline
\multicolumn{1}{|l|}{BERT} & \multicolumn{1}{l|}{0.531} & 0.322 \\ \hline
\multicolumn{1}{|l|}{RoBERTa} & \multicolumn{1}{l|}{0.495} & 0.431 \\ \hline
\multicolumn{1}{|l|}{XLM-R} & \multicolumn{1}{l|}{0.488} & 0.422 \\ \hline
\multicolumn{1}{|l|}{XLM-T} & \multicolumn{1}{l|}{0.494} & 0.534 \\ \hline
\multicolumn{3}{|c|}{\textit{Original + translated}} \\ \hline
\multicolumn{1}{|l|}{BERT} & \multicolumn{1}{l|}{0.545} & 0.353 \\ \hline
\multicolumn{1}{|l|}{XLM-R} & \multicolumn{1}{l|}{0.504} & 0.467 \\ \hline
\multicolumn{1}{|l|}{XLM-T} & \multicolumn{1}{l|}{0.455} & 0.554 \\ \hline
\end{tabular}
\caption{\label{font-table} Development phase. Evaluation on the validation subset.}
\label{table:dev_all}
\end{table}

\begin{figure}
\begin{minipage}[h]{1\linewidth}
\center{\includegraphics[width=0.8\linewidth]{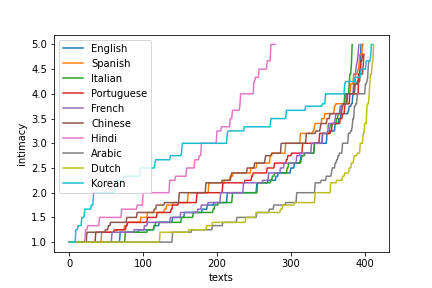} \\ a) gold labels}
\end{minipage}
\vfill
\begin{minipage}[h]{1\linewidth}
\center{\includegraphics[width=0.8\linewidth]{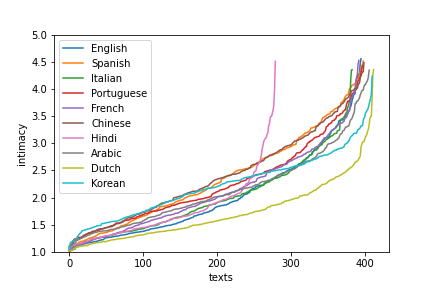} \\ b) predictions}
\end{minipage}
\caption{The distribution of intimacy scores for each language (test set).}
\label{ris:image1}
\end{figure}

Further, we evaluated the effectiveness of different input representations for zero-shot tweet intimacy prediction. For this purpose, we excluded languages from the original training set one by one. The models were fine-tuned on the remaining languages and then evaluated on the excluded language. We did not exclude English since we utilized English as an auxiliary language during the formation of input representations. The results are demonstrated in Table \ref{table:dev_unseen}. The first column presents the types of input representation for training and validation subsets. For example, the first row shows the results obtained for the models fine-tuned and evaluated on the original data.

The lowest Pearson's r scores for the original dataset were obtained for Chinese (0.153), French (0.158), and Italian (0.162). The result for the Portuguese language was more competitive. For Spanish, we achieved 0.579 which is higher than the score obtained on the mixed language validation set (0.572) in Table \ref{table:dev_all}. Fine-tuning on data translated into English as well as fine-tuning on the original dataset and the subsequent validation on translated data in most cases led to a deterioration in the results. However, a significant growth in Pearson's was obtained for Chinese while using the translated training set (+0.23). Slight improvements were also demonstrated for Italian (+0.02 of Pearson's r) and French (+0.069) for the model fine-tuned on translated texts and validated on the original dataset. The best MSE scores for Italian and Chinese were shown on translated data. Taking into account the relatively low Pearson's scores, it can be assumed that the model predicts close to the average values for any data. The joint fine-tuning using original and translated data was superior compared to the fine-tuning on the original dataset for three of the five languages (Italian, French, and Chinese) in terms of Pearson's r. Moreover, the average Pearson's r was 0.301 for original data and 0.426 for the joint use of original and translated texts. The growth of the averaged Pearson's r using translated texts as auxiliary data was 0.125 in our experiments.

\begin{table*}[]
\centering
\small
\begin{tabular}{|l|l|l|l|l|l|l|l|l|l|l|}
\hline
\multicolumn{1}{|l|}{\multirow{3}{*}{Model}} & \multicolumn{10}{c|}{Unseen language} \\ \cline{2-11} 
\multicolumn{1}{|l|}{} & \multicolumn{2}{l|}{Spanish} & \multicolumn{2}{l|}{Italian} & \multicolumn{2}{l|}{Portuguese} & \multicolumn{2}{l|}{French} & \multicolumn{2}{l|}{Chinese} \\ \cline{2-11} 
\multicolumn{1}{|l|}{} & \multicolumn{1}{l|}{\textit{MSE}} & \multicolumn{1}{l|}{\textit{r}} & \multicolumn{1}{l|}{\textit{MSE}} & \multicolumn{1}{l|}{\textit{r}} & \multicolumn{1}{l|}{\textit{MSE}} & \multicolumn{1}{l|}{\textit{r}} & \multicolumn{1}{l|}{\textit{MSE}} & \multicolumn{1}{l|}{\textit{r}} & \multicolumn{1}{l|}{\textit{MSE}} & \multicolumn{1}{l|}{\textit{r}} \\ \hline
\begin{tabular}[c]{@{}l@{}}Train: original\\ Validation: original\end{tabular} & \textbf{0.414} & \textbf{0.579} & 0.452 & 0.162 & 0.469 & \textbf{0.452} & \textbf{0.435} & 0.158 & 0.587 & 0.153 \\\hline
\begin{tabular}[c]{@{}l@{}}Train: original\\ Validation: translated\end{tabular} & 0.513 & 0.205 & 0.463 & 0.142 & 0.46 & 0.227 & 0.51 & 0.101 & 0.677 & 0.113 \\\hline
\begin{tabular}[c]{@{}l@{}}Train: translated\\ Validation: translated\end{tabular} & 0.461 & 0.103 & 0.451 & 0.11 & \textbf{0.455} & 0.134 & 0.483 & 0.123 & \textbf{0.541} & 0.383 \\\hline
\begin{tabular}[c]{@{}l@{}}Train: translated\\ Validation: original\end{tabular} & 0.465 & 0.444 & 0.471 & 0.182 & 0.532 & 0.19 & 0.508 & 0.227 & 0.599 & 0.383 \\\hline
\begin{tabular}[c]{@{}l@{}}Train, validation:\\ original + translated\end{tabular} & 0.42 & 0.565 & \textbf{0.436} & \textbf{0.259} & 0.462 & 0.439 & 0.459 & \textbf{0.32} & 0.577 & \textbf{0.547}\\\hline
\end{tabular}
\caption{\label{font-table} Development phase. Evaluation on unseen languages.}
\label{table:dev_unseen}
\end{table*}

On the basis of our experiments, we combined the XLM-T fine-tuned on the original dataset provided by the organizers of the competition with the XLM-T fine-tuned in a joint manner simultaneously utilizing original and translated texts. The first type of model was used for making predictions on the seen languages (English, Spanish, Italian, Portuguese, French, and Chinese) while the second type was applied to the unseen languages (Hindi, Dutch, Korean, and Arabic).

\subsection{Overall Performance}

Similar to our previous work on tweet analysis \cite{glazkova2021g2tmn,glazkova2021fine}, we applied ensemble learning for the final submission. We used a two-part approach based on the ensembles of the XLM-T fine-tuned on the original and joint data for the seen and unseen languages respectively (Figure \ref{fig}). We fine-tuned seven models on each data type and took an average across all the models of the same type.

Table \ref{table:test_results} shows the official scores of our team. The proposed model ranked 4th at SemEval-2023 Task 9 and obtained an overall Pearson's r of 0.599 over the test set. This result improved up to 0.088 Pearson's r over a score averaged across all 45 submissions. The proposed model showed above-average scores for all languages from the dataset and entered the top ten teams for both seen and unseen languages. The highest results were obtained for Arabic (2nd), Hindi (5th), Italian (6th), and French (9th).

\begin{table*}[]
\centering
\small
\begin{tabular}{|l|l|l|l|l|l|l|l|l|l|l|l|l|l|}
\hline
 & \multicolumn{13}{c|}{Language} \\ \cline{2-14}
\multirow{-2}{*}{Score} & \multicolumn{1}{l|}{\begin{sideways}English\end{sideways}} & \multicolumn{1}{c|}{\begin{sideways}Spanish\end{sideways}} & \multicolumn{1}{c|}{\begin{sideways}Portuguese\end{sideways}} & \multicolumn{1}{c|}{\begin{sideways}Italian\end{sideways}} & \multicolumn{1}{c|}{\begin{sideways}French\end{sideways}} & \multicolumn{1}{c|}{\begin{sideways}Chinese\end{sideways}} & \multicolumn{1}{c|}{\begin{sideways}Hindi\end{sideways}} & \multicolumn{1}{c|}{\begin{sideways}Dutch\end{sideways}} & \multicolumn{1}{c|}{\begin{sideways}Korean\end{sideways}} & \multicolumn{1}{c|}{\begin{sideways}Arabic\end{sideways}} & \multicolumn{1}{c|}{\begin{sideways}Seen\end{sideways}} & \multicolumn{1}{c|}{\begin{sideways}Unseen\end{sideways}}& \begin{sideways}Overall\end{sideways} \\ \hline
\textit{r} & 0.717 & 0.74 & 0.684 & 0.734 & 0.708 & 0.721 & 0.242 & 0.639 & 0.361 & 0.662 & 0.637 & 0.357&0.599 \\ 
Rank & 11 & 10 & 12 & 6 & 9 & 15 & 5 & 9 & 17 & 2 & 9&9&4 \\ \hline
avg \textit{r}& 0.635 & 0.664 & 0.596 & 0.636 & 0.611 & 0.629 & 0.187 & 0.539 & 0.407 & 0.524 & 0.727 & 0.441& 0.511 \\ \hline
\end{tabular}
\caption{\label{font-table} Official results.}
\label{table:test_results}
\end{table*}

\subsection{Error Analysis}

Figure \ref{ris:image1} shows the distribution of intimacy scores in the official test set and in the predictions made using our approach. From the chart, it can be seen that the maximum values predicted by our model are lower than in the official set. The maximum result predicted by the model is 4.56 and the minimum result is 0.99 while test data contains values between 1 and 5, inclusive. The scores produced by the model are closer to the average scores. The average value for the official scores is 2.13 with a standard deviation of 0.97. For our predictions, the average value is 2.11 and the standard deviation is 0.74.

\begin{figure}
\begin{minipage}[h]{0.49\linewidth}
\center{\includegraphics[width=0.64\linewidth]{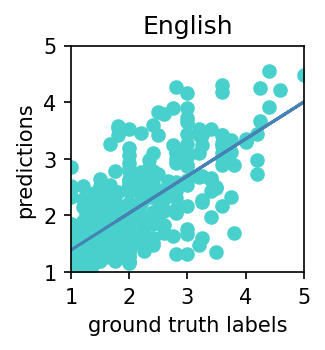}}  \\
\end{minipage}
\hfill
\begin{minipage}[h]{0.49\linewidth}
\center{\includegraphics[width=0.64\linewidth]{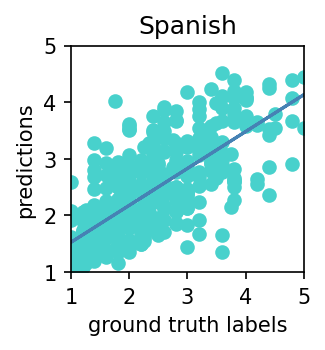}} \\
\end{minipage}
\vfill
\begin{minipage}[h]{0.49\linewidth}
\center{\includegraphics[width=0.64\linewidth]{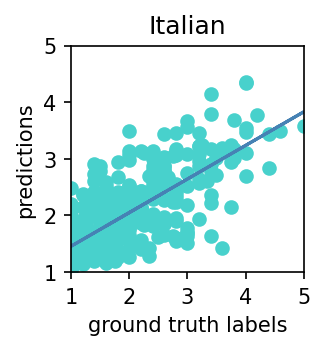}} \\
\end{minipage}
\hfill
\begin{minipage}[h]{0.49\linewidth}
\center{\includegraphics[width=0.64\linewidth]{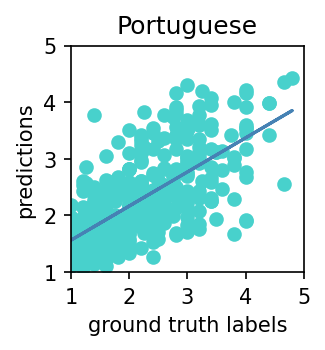}}  \\
\end{minipage}
\vfill
\begin{minipage}[h]{0.49\linewidth}
\center{\includegraphics[width=0.64\linewidth]{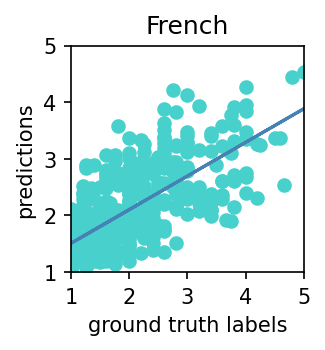}} \\
\end{minipage}
\hfill
\begin{minipage}[h]{0.49\linewidth}
\center{\includegraphics[width=0.64\linewidth]{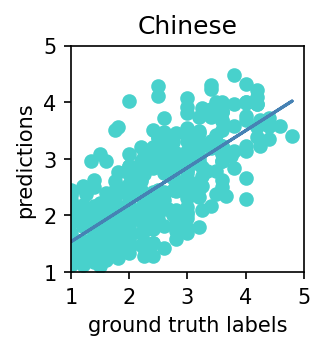}}  \\
\end{minipage}
\vfill
\begin{minipage}[h]{0.49\linewidth}
\center{\includegraphics[width=0.64\linewidth]{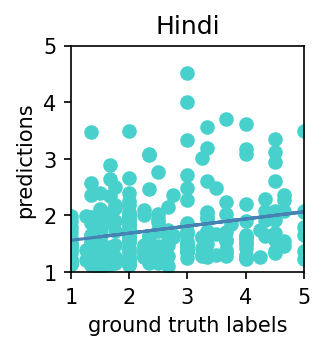}} \\
\end{minipage}
\hfill
\begin{minipage}[h]{0.49\linewidth}
\center{\includegraphics[width=0.64\linewidth]{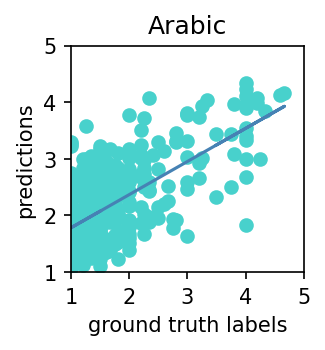}} \\
\end{minipage}
\vfill
\begin{minipage}[h]{0.49\linewidth}
\center{\includegraphics[width=0.64\linewidth]{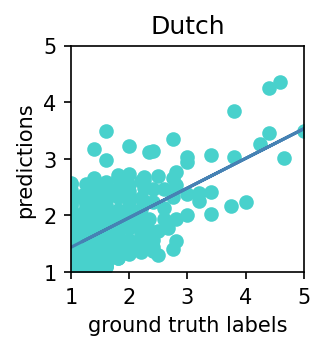}} \\
\end{minipage}
\hfill
\begin{minipage}[h]{0.49\linewidth}
\center{\includegraphics[width=0.64\linewidth]{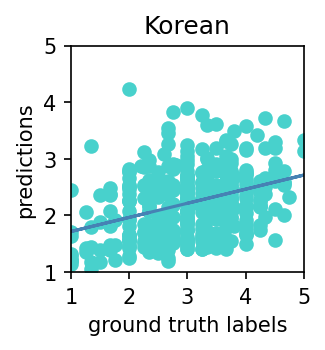}} \\
\end{minipage}
\caption{The correlation between predictions and ground truth test labels.}
\label{ris:correlation}
\end{figure}

The correlation between the predictions and ground truth test labels is shown in Figure \ref{ris:correlation}. In general, the results for the seen languages are predictably better. The lowest correlation among all languages was obtained for Hindi and Korean.

\section{Conclusion}\label{sec6}
In this work, we have explored an application of pre-trained language models and translated data for the task of multilingual tweet intimacy detection. We showed that the joint use of original texts and texts translated into English improves the zero-shot performance of XLM-T for unseen languages. Further research in zero-shot intimacy prediction could be conducted to estimate the quality of the translations used and the assessment of the proximity of languages from training and test sets.

\bibliography{semeval23-clickbait-spoiling-system-paper}
\bibliographystyle{acl_natbib}
\newpage
\appendix
\section{Appendix}\label{appendix}

\begin{table}[hbt]
\centering
\small
\begin{tabular}{|p{5.7cm}|p{1cm}|}
\hline
Tweet & \multicolumn{1}{l|}{Intimacy} \\ \hline
Here is a nice equation: 0+0-0-0+0=0 & 1.0 \\ \hline
Try that again Timmy... Ill turn you into bonemeal http & 1.4 \\ \hline
@user @DoorDash Damn Shame, you tipped a \$1.30 & 2.25 \\ \hline
@user @verge I can't believe you don't value family & 3.0 \\ \hline
i’ll be yours through all the years, till the end of time. \#MewSuppasit @MSuppasit \#MyCandyHeroxMew & 4.0 \\ \hline
fren: who you smiling about me: nothing, just the love of my life the love of my life: http & 4.4 \\ \hline
\end{tabular}
\caption{A sample of tweets from the dataset.}
\end{table}

\end{document}